\documentclass[conference]{IEEEtran}
\usepackage{times}

\usepackage[numbers]{natbib}
\usepackage{multicol}
\usepackage[bookmarks=true]{hyperref}

\usepackage{amssymb}
\usepackage{amsmath}
\usepackage{float}
\usepackage{dblfloatfix}
\usepackage{graphicx}
\usepackage{mathtools}
\usepackage{siunitx}
\usepackage{filecontents}
\usepackage[font=footnotesize,labelfont=footnotesize]{subcaption}
\usepackage[font=footnotesize,labelfont=footnotesize]{caption}

\usepackage{comment}
\usepackage{multirow}
\usepackage{environ}

\usepackage{amsthm}

\usepackage{algorithm, algorithmicx}
\usepackage[noend]{algpseudocode}

\usepackage{xcolor}

\algnewcommand\algorithmicinput{\textbf{Input:}}
\algnewcommand\algorithmicoutput{\textbf{Output:}}
\algnewcommand\algorithmicinitialize{\textbf{Initialize:}}
\algnewcommand\Input{\item[\algorithmicinput]}%
\algnewcommand\Output{\item[\algorithmicoutput]}%
\algnewcommand\Initialize{\item[\algorithmicinitialize]}%

\graphicspath{ {img/} }
\begin{document}

\title{Reconfigurable Robot Control Using Flexible Coupling Mechanisms}

\author{\authorblockN{Sha Yi, Katia Sycara, and Zeynep Temel}
\authorblockA{Robotics Institute\\
Carnegie Mellon University, Pittsburgh, USA\\
Email: {\tt\small \{shayi, katia, ztemel\}@cs.cmu.edu}.}}

\maketitle
\begin{abstract}
Reconfigurable robot swarms are capable of connecting with each other to form complex structures. Current mechanical or magnetic connection mechanisms can be complicated to manufacture, consume high power, have a limited load-bearing capacity, or can only form rigid structures. In this paper, we present our low-cost soft anchor design that enables flexible coupling and decoupling between robots. Our asymmetric anchor requires minimal force to be pushed into the opening of another robot while having a strong pulling force so that the connection between robots can be secured. To maintain this flexible coupling mechanism as an assembled structure, we present our Model Predictive Control (MPC) frameworks with polygon constraints to model the geometric relationship between robots. We conducted experiments on the soft anchor to obtain its force profile, which informed the three-bar linkage model of the anchor in the simulations. We show that the proposed mechanism and MPC frameworks enable the robots to couple, decouple, and perform various behaviors in both the simulation environment and hardware platform. Our code is available at \url{https://github.com/ZoomLabCMU/puzzlebot_anchor}. Video is available at \url{https://www.youtube.com/watch?v=R3gFplorCJg}.
\end{abstract}
\IEEEpeerreviewmaketitle

\section{Introduction}
Robot swarms demonstrated collective behaviors that a single robot could not accomplish \cite{nagavalli2017automated, pickem2017robotarium, werfel2014designing}. Furthermore, robots that physically connect and interact with each other, i.e., the self-reconfigurable robots, have extended capabilities and behaviors. Applications include reconfiguration for terrain adaptation \cite{ion2015adaptation}, dynamic infrastructure \cite{mateos2019autonomous}, collaborative transportation \cite{saldana2018modquad, alonso2017multi}, search and rescue \cite{whitman2018snake}. Our goal is to develop a robot swarm system where robots can dynamically couple to form functional structures, e.g., bridges and ropes, and decouple to navigate around the environment individually. 

Many modular robots that can dynamically couple and decouple use magnetic forces, e.g., ATRON \cite{jorgensen2004modular}, M-TRAN III \cite{kurokawa2008distributed}, and M-blocks \cite{romanishin20153D}. Since the power and dynamic structures are dedicated to the coupling mechanism, each module has limited mobility compared with a regular mobile robot, which limits their ability to navigate around the environment. 

To be able to move around efficiently, the robots need efficient locomotion approaches, e.g., wheels or legs. Swarm-bot \cite{gross2006autonomous} are wheeled robots with grippers that can grip onto each other to form bridges. 
The SMORES \cite{tosun2016Design} and FreeBot\cite{liang2020freebot} modules are equipped with wheels and can form 3D structures with magnetic forces. However, these \textit{actuated} coupling mechanisms consume high power when maintaining the coupling status. Moreover, although magnetic connections provide high pulling forces, the shear force is significantly lower. The polarity of magnets also limits the connection to be directional, and the strength of the connection is highly sensitive to misalignment \cite{tosun2016Design}. Compared with the \textit{active} actuated coupling mechanisms, \textit{passive} coupling strategies - where no power is dedicated to maintaining the coupling status - are more energy efficient. The ModQuad\cite{saldana2018modquad}, Roboats \cite{mateos2019autonomous}, SlimeBot \cite{shimizu2009amoeboid}, and the PuzzleBots \cite{yi2021puzzlebots} utilize passive coupling mechanisms to form and maintain structures. However, with these passive mechanisms, robots form a rigid 2D connection when coupled together, which greatly limits their functionality and mobility.

\begin{figure}
    \centering
    \includegraphics[width=0.47\textwidth]{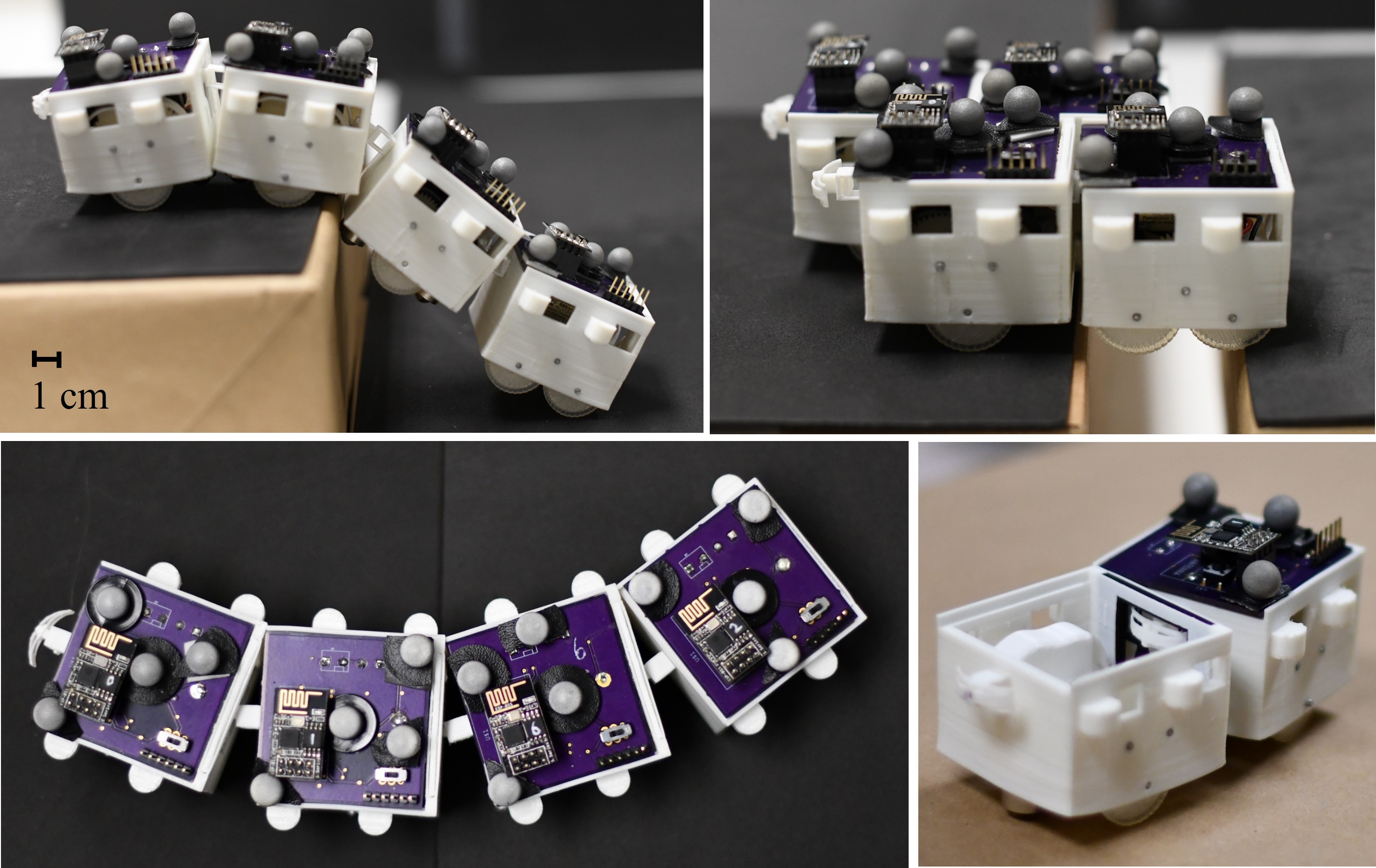}
    \caption{Top: Robots can form a flexible chain and go down a step (left) or a rigid structure that crosses a gap between two equal-height test stages (right). Bottom left: Four robots forming an articulated chain. Bottom right: One robot inserts its anchor inside the opening of another one (circuitry removed for visual purposes).}
    \label{fig:intro_demo}
    \vspace{-5pt}
\end{figure}

Although passive coupling mechanisms have the benefits of power efficiency, maintaining the coupled states require precise control of the robot's motion. 
The PuzzleBots \cite{yi2022configuration} utilize quadratic programming (QP) based configuration control algorithm that constrains a pair of points between two robots to maintain the coupled state. This point constraint is very sensitive to localization errors and heavily restricts the robot's motion. Control Barrier Functions are commonly used to relax distance-based constraints \cite{zeng2021safety, borrmann2015control}. However, considering robots as point masses and their interactions as point pairs does not fully capture the geometry of the coupling behaviors. Instead, polytopes are a more precise option to describe geometries. Most studies that focus on the geometry of robots and objects are for collision avoidance \cite{thirugnanam2022safety, thirugnanam2022duality}. However, to maintain the connection of passive coupling mechanisms, robots need to keep their connection points within polygon shapes instead of avoiding collision with polytopes.

Our goal is to design a passive coupling mechanism that enables robots to 
form functional configurations that are compliant with environment structures, and control the assembled configuration to perform collective behaviors. To be compliant with unstructured environments and traverse terrain with large vertical drops,
the coupling mechanism should provide strong enough force to hold the load of multiple robots. While having a strong connection, the robots should still be able to decouple from each other dynamically. It is challenging to balance a strong connection and an easy disconnecting method on a single passive mechanism with no dedicated power. The assembly formed by multiple robots should also be flexible in terms of mobility so that the assembled structure can move in a desirable configuration forming a bridge over a gap or a rope down a stair. Modeling and controlling such a flexible structure is also challenging. Soft objects are difficult to model due to their deformability and the lack of state and force estimation. With a passive coupling mechanism, the robots need precise control over their motion to maintain the connection. Such maintenance constraints should be restrictive enough so that the robots do not decouple easily and relaxed enough so that the assembled structure can move around the environment.

In this paper, we propose a \textit{soft anchor} design that is passive and has a strong holding force for 
vertical structures
while keeping a simple coupling and decoupling behavior between robots. The anchor acts as an unactuated joint when inserted into the opening of another robot. The anchor is asymmetric - the force of pushing it into an opening is small, while the force required to pull it out is large. 
This unactuated joint formed by the soft anchor also leaves room for robots in an assembled structure to rotate or translate within the limit of the joint, compliant with the robot's motion. We limit our soft anchor design to having only one degree of freedom (yaw rotation), and its state can be estimated by the relative position and displacement. By obtaining force profiles for our soft anchor, we can model and simulate it as a linkage of rigid components similar to \cite{zheng2022scalable}. We maintain the connection between robots by enforcing a polygon-based constraint. This restricts the relative motion between robots to maintain the coupling status while giving some flexibility to the robots to perform rotations or translations. 


Our contribution lies in three aspects. First, we propose the design of a soft asymmetric anchoring mechanism that allows robots to couple and form flexible or rigid configurations. Second, we present a three-bar linkage system in simulation that models the soft anchor and simulate robot behaviors comparable to the actual hardware system. Finally, based on our designed mechanism, we propose a Model Predictive Control (MPC) based framework with polygon constraints to model the precise geometry of the coupling formations. 

\section{Design of Flexible Coupling Mechanisms}
In this section, we will first review our robot design. Similar to \citet{yi2022configuration}, we have \textit{pilot} robots with four actuated wheels for better ground grip and \textit{non-pilot} robots with two actuated wheels to perform agile motions for coupling and decoupling. Based on the robot model, we propose our flexible passive anchor design that provides coupling and decoupling behavior between robots, as well as a locking behavior for maintaining configurations on unstructured terrain.

\subsection{Robot Design}
The robots, shown in Figure~\ref{fig:intro_demo}, are 50 mm in depth (excluding the anchor), 50 mm in width, 45 mm in height (without WiFi and optical markers), and weighs 68 g. There are four markers on the top of the robot for the motion-tracking system (Vicon) to provide high-frequency pose information. Each robot has an ESP8266 Wifi Module that connects to the same network with the Vicon and a central computer. The onboard power is provided with a 3V CR2 battery. Two DC motors execute velocity commands received from the central computer. 
Design configurations to utilize the new anchor structure are explained in the next section.

\subsection{Anchor Design}

\begin{figure*}
\centering
\begin{subfigure}{0.2\textwidth}
\centering
\includegraphics[width=\textwidth]{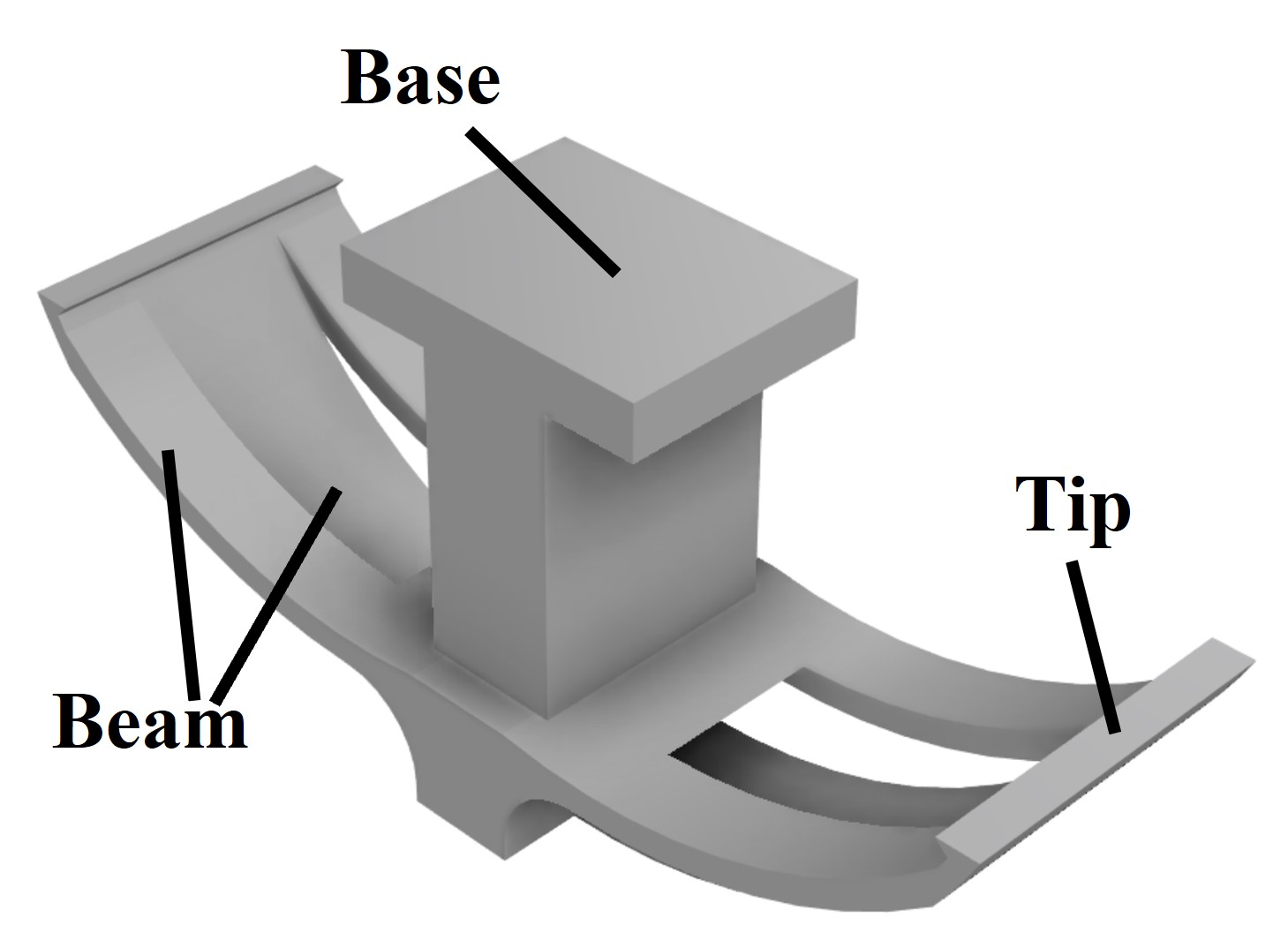}
\caption{}
\label{fig:fusion_home}
\end{subfigure}\hfill
\begin{subfigure}{0.2\textwidth}
\centering
\includegraphics[width=\textwidth]{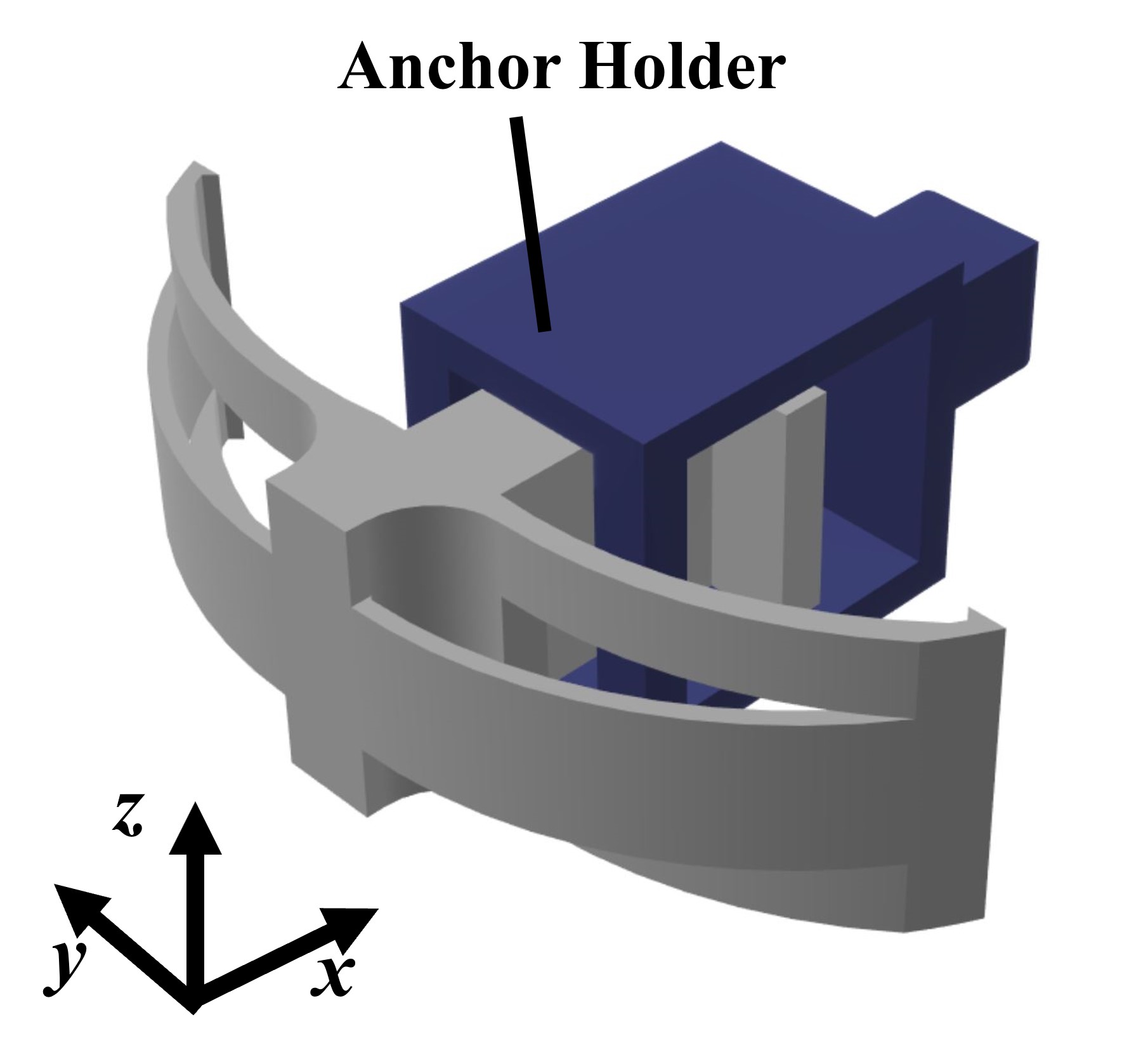}
\caption{}
\label{fig:anchor_holder}
\end{subfigure}\hfill
\begin{subfigure}{0.16\textwidth}
\centering
\includegraphics[width=\textwidth]{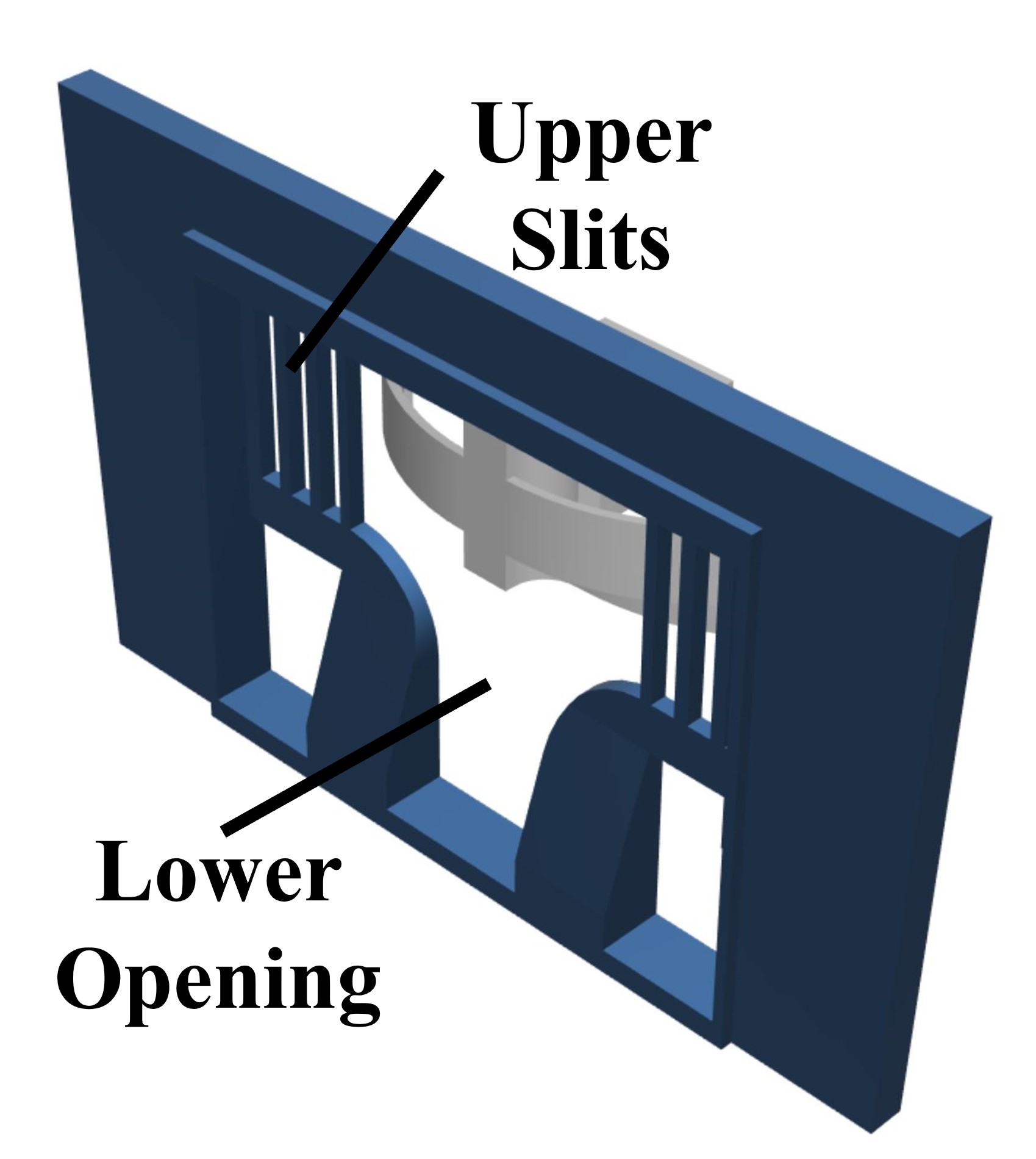}
\caption{}
\label{fig:assembly_out}
\end{subfigure}\hfill
\begin{subfigure}{0.16\textwidth}
\centering
\includegraphics[width=\textwidth]{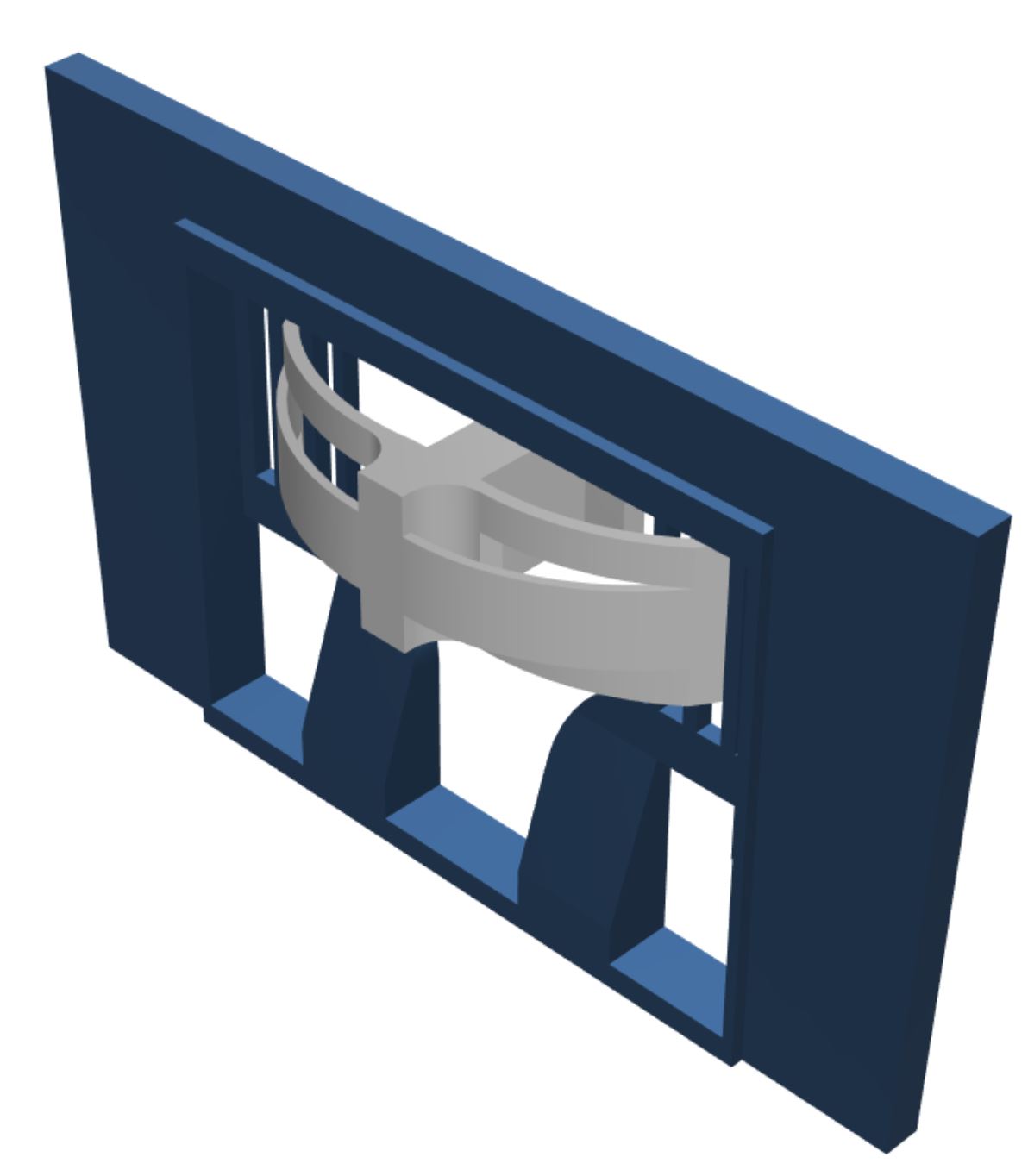}
\caption{}
\label{fig:assembly_in}
\end{subfigure}\hfill
\begin{subfigure}{0.16\textwidth}
\centering
\includegraphics[width=\textwidth]{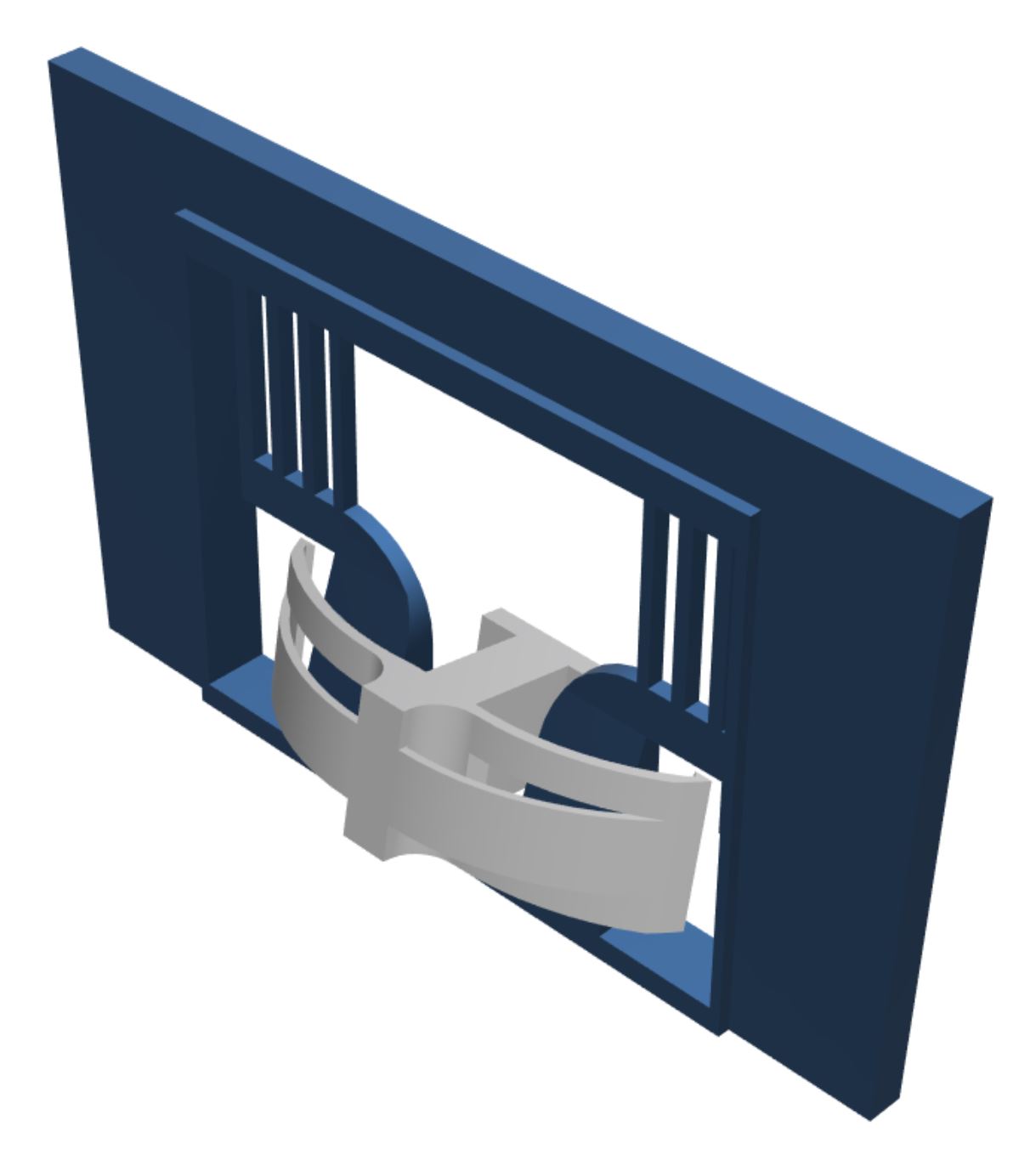}
\caption{}
\label{fig:assembly_lock}
\end{subfigure}
\caption{(a) Anchor design. (b) Anchor (gray) with its base inserted in a holder (blue) acting as a floating joint. (c) Anchor and an opening uncoupled when on the same height. (d) Anchor and opening coupled. The anchor tips are sitting in the slits. (e) Anchor locks the connection when the robot's vertical position changes with gravity due to the irregularities on the ground. }
\vspace{-1pt}
\label{fig:anchor_design}
\end{figure*}

\begin{figure*}
\centering
\begin{subfigure}{0.32\textwidth}
\includegraphics[width=\textwidth]{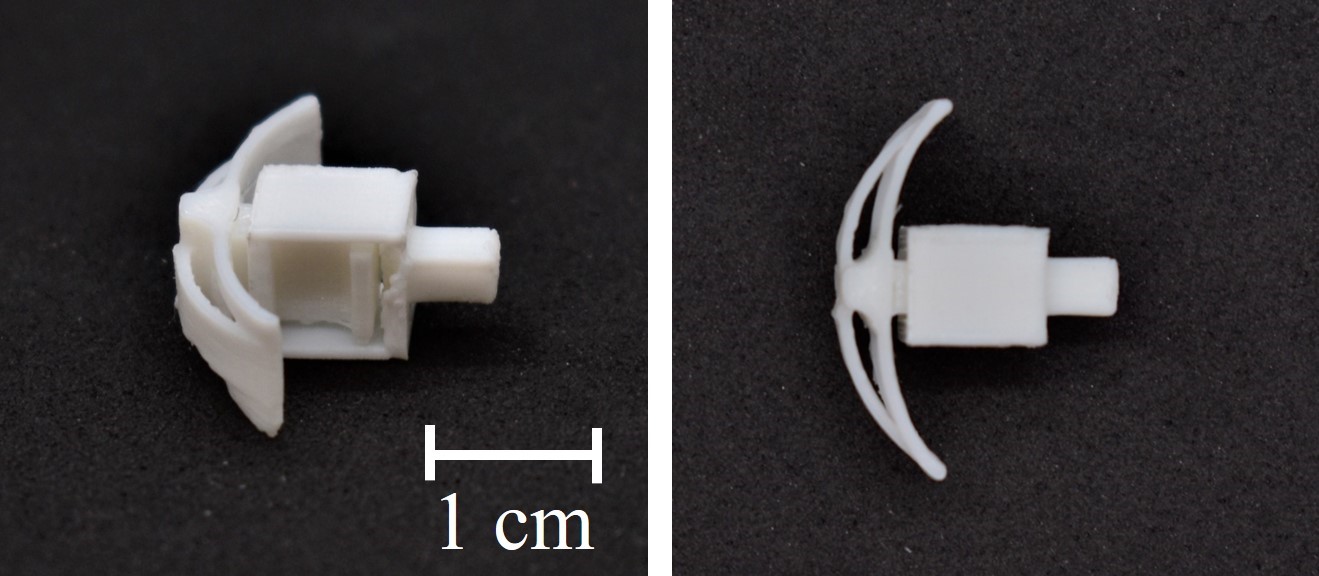}
\caption{} \label{fig:anchor_rest}
\end{subfigure}\hfill
\begin{subfigure}{0.32\textwidth}
\includegraphics[width=\textwidth]{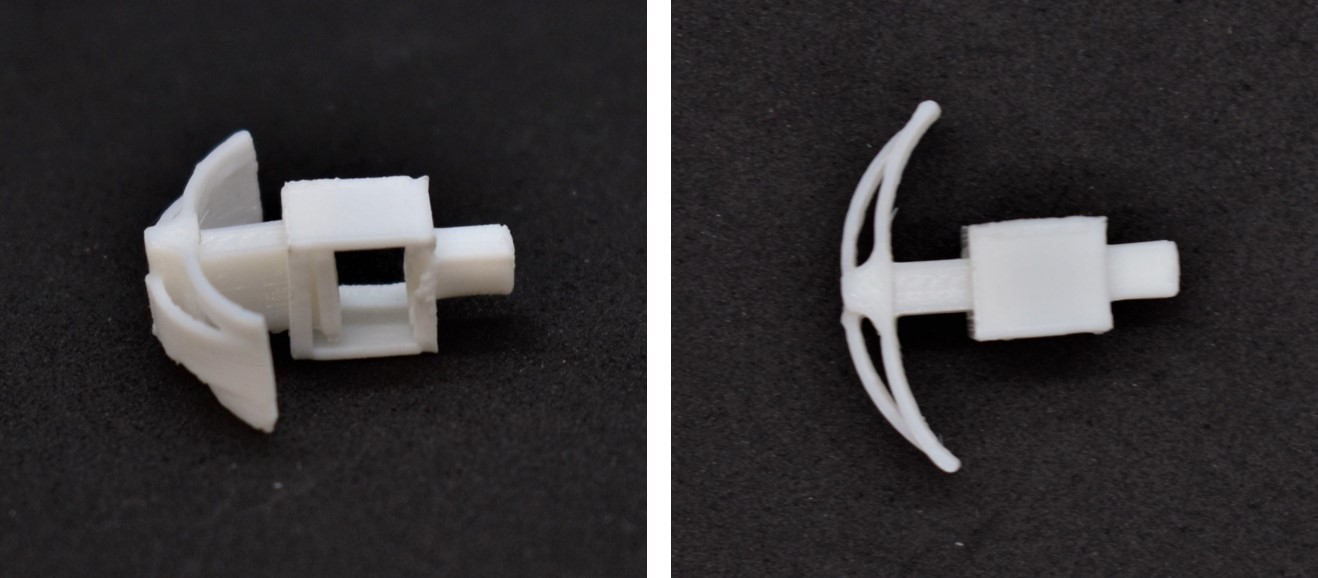}
\caption{} \label{fig:anchor_extend}
\end{subfigure}\hfill
\begin{subfigure}{0.32\textwidth}
\includegraphics[width=\textwidth]{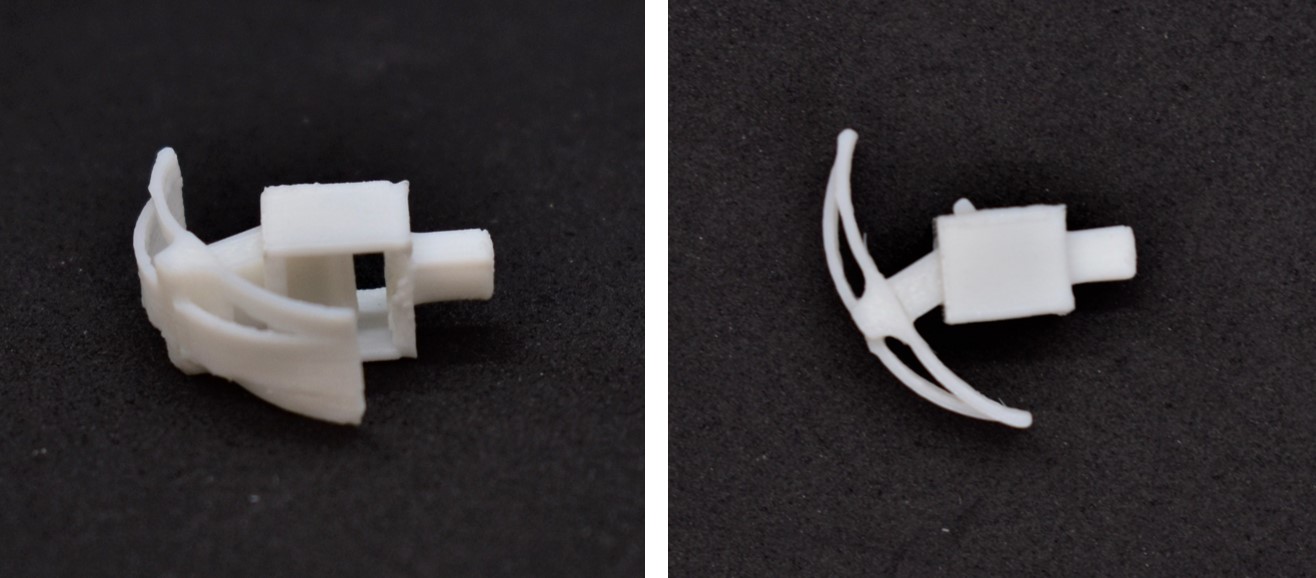}
\caption{} \label{fig:anchor_rotate}
\end{subfigure}
\caption{States of the 3D printed anchor from two viewpoints. (a) Anchor collapsed. In this state, the anchor is in its shortest configuration. (b) Anchor extended. The holder provides translational motion within 5 mm. (c) Anchor rotated. When not collapsed, the anchor is also free to perform rotations within 0.5 rad.} 
\label{fig:anchor_photos}
\vspace{-1pt}
\end{figure*}

Existing flexible coupling mechanisms that utilize magnets have limited shear force \cite{tosun2016Design, saldana2018modquad, liang2020freebot}, active mechanical connections are power-consuming and complicated to manufacture and integrate with a different platform \cite{gross2006autonomous, mateos2019autonomous}, while the fully passive connections are rigid and not robust with disturbances \cite{yi2021puzzlebots}. Therefore, the design goal of our coupling mechanism includes 1) strong enough forces to hold multiple robots at the same time; 2) a fully passive coupling mechanism so that no additional power is consumed for the coupling process; 3) flexible as an assembled structure while being robust on rough terrains and with uncertainties from actuation and localization. 

With the above-mentioned requirements, we propose our anchor coupling design shown in Figure~\ref{fig:anchor_design}. Figure~\ref{fig:fusion_home} shows the anchor with two shorter beams on the side and one longer beam in the middle. The beams are curved towards the base of the anchor. The anchor is 3D printed with Thermoplastic Polyurethane (TPU) (NinjaFlex Cheetah). With this design, the forward pushing force of the anchor can be small, and the backward pushing force of the anchor can be large. Characterization and force profiles are presented in Section~\ref{sec:characterization}. As shown in Figure~\ref{fig:anchor_holder}, the anchor sits in a rigid base, which is 3D printed with a hard plastic material, Polylactic Acid (Ultimaker Tough PLA). The soft anchor (gray) is placed inside the anchor holder (blue), through which the anchor is connected to the robot's body.
The anchor can rotate freely in the yaw direction and moves freely along the $x$ axis inside the holder. The anchor has limited motion in the $y$ direction ($< 0.5$ mm) or pitch rotation ($< 0.1$ rad). This floating joint between the anchor and the holder gives flexibility when the robots are coupled together. The anchor and its holder are mounted on the back of the robot, and there is an opening in the front of the robot shown in blue as in Figure~\ref{fig:assembly_out}, which is a state when two robots are not coupled. As shown in Figure~\ref{fig:assembly_in}, one robot can push its anchor inside the opening of another robot to couple. The anchor tips will sit in the slits on the two sides of the opening. With the anchor beam design we introduced, the force of pushing the anchor inside will be significantly smaller than the force provided by the root during forward motion. The force needed to pull the anchor backward is more significant than the wheel actuator force. Thus, the anchor will be easy to push inside the opening for the robots to couple and cannot decouple easily. The robot needs to wiggle to decouple from each other. We show this decoupling controller in Section~\ref{sec:decouple}. By adding different constraints over the robot's motion, we may avoid the wiggling behavior when we would like the robots to couple or stay coupled, and only enable the wiggling behavior if we want the robots to decouple. This provides robustness under uncertainty with actuation and localization in the hardware system. This motion is presented in Section~\ref{sec:result}. When one robot moves toward an edge of a test stage, it will drop with gravity. The anchor will drop with the robot and sits in the slot as shown in Figure~\ref{fig:assembly_lock}. This locks the robot's movement, and the assembly can form a chain down the stage's edge. This anchor-opening connection type provides additional flexibility for the coupled assembly. The robots can still couple on the side with the knob-hole connection as in \cite{yi2022configuration}, which can be necessary when reaching a test stage of the same height across a gap.

In summary, the geometry and soft material of the anchor enables flexible coupling and decoupling behavior while providing robustness during execution. We show the hardware experiments in Section~\ref{sec:hardware} and in our supplementary video material. To achieve this on our non-linear differential drive platform, we will introduce our multi-robot Model Predictive Control (MPC) method with polygon-based constraints in the next section.

\section{Modeling and Control of Coupling Behaviors}\label{sec:model_and_control}
To maintain the coupling status, one way is to maintain a minimal distance between two specific points (connection points) on the robot bodies. This is very restrictive on the robot's motion and not robust against noise or disturbances. In this work, we incorporate the anchor coupling mechanism introduced in the previous section on a unicycle dynamics model. To this end, we utilize Model Predictive Control (MPC) in our multi-robot system along with polygon constraints to maintain the coupling status. In this section, we introduce the basic setup for MPC and the sequential coupling behavior for two robots. Then we propose the polygon-based constraint for maintaining the coupling pairs in an assembly. The actual hardware system consists of a low-frequency central MPC planner and a high-frequency PID controller to be introduced in Section~\ref{sec:hardware}.

\subsection{Robot Dynamics}\label{sec:robot_dynamics}
Consider a multi-robot system with $N$ robots on a 2D plane. The state of robot $i$, where $i \in \{1, 2, \dots, N\}$, is $\mathbf{x}_i = [p_x, p_y, \theta, v, w]^{\intercal}$. $p_x, p_y \in \mathbb{R}$ are the position in the spacial frame $x$ and $y$ axis respectively. $\theta \in (-\pi, \pi]$ is the heading angle (yaw) of the robot. The $v$ and $w$ are linear and angular velocities, respectively. Control input for robot $i$ is $\mathbf{u}_i = [\dot v_i, \dot w_i]^{\intercal}$. Consider robot dynamics as follows (we temporarily drop the robot index $i$ to simplify notations)
\begin{equation}\label{eq:single_dynamics}
\dot{\mathbf{x}} = f(\mathbf{x}, \mathbf{u}) 
= \begin{bmatrix}
v\, \cos \theta \\
v\, \sin \theta \\
w \\
0 \\ 0
\end{bmatrix} 
+ \begin{bmatrix}
0 & 0 \\ 0 & 0 \\ 0 & 0 \\ 1 & 0 \\ 0 & 1
\end{bmatrix} \mathbf{u} 
\end{equation}
In Section~\ref{sec:mpc_setup}, we integrate this dynamic equation with the Euler method for discretization in the MPC setup. Although there are more accurate discretization methods available like RK4 \cite{atkinson1991introduction}, due to our short MPC horizon, there is no noticeable difference between the Euler and RK4 discretization in our experiments. Thus, we choose the Euler discretization method for faster computation.

\subsection{Decoupling Behavior} \label{sec:decouple}
We introduce the decoupling behavior first due to its simplicity. When two robots are coupled, the anchor of one robot sits inside the opening of another robot, as shown in Figure~\ref{fig:assembly_in}. To decouple, the robot with the anchor wiggles when the anchor is in the upper slits, i.e., periodically rotates and moves with a linear velocity. This wiggling behavior is controlled as
\begin{equation}\label{eq:wiggle}
\begin{bmatrix}
    v \\ w
\end{bmatrix} = \begin{bmatrix}
v_{\text{bias}} \\ w_{\max}\sin(B t)
\end{bmatrix}
\end{equation}
where $2\pi/B$ is the period. The rotation from this behavior loosens the anchor tip from the upper slits of the opening and makes it free to rotate inside the robot's body. With the compliance in the anchor tip, this rotation in the anchor results in the decoupling of robots from each other. 
We show this behavior in Section~\ref{sec:hardware}. 

\subsection{Connection as Polygon Constraints}
\subsubsection{Projected Anchor Zero Position}

\begin{figure*}
\centering
\begin{subfigure}{0.3\textwidth}
    \centering
    \includegraphics[width=\textwidth]{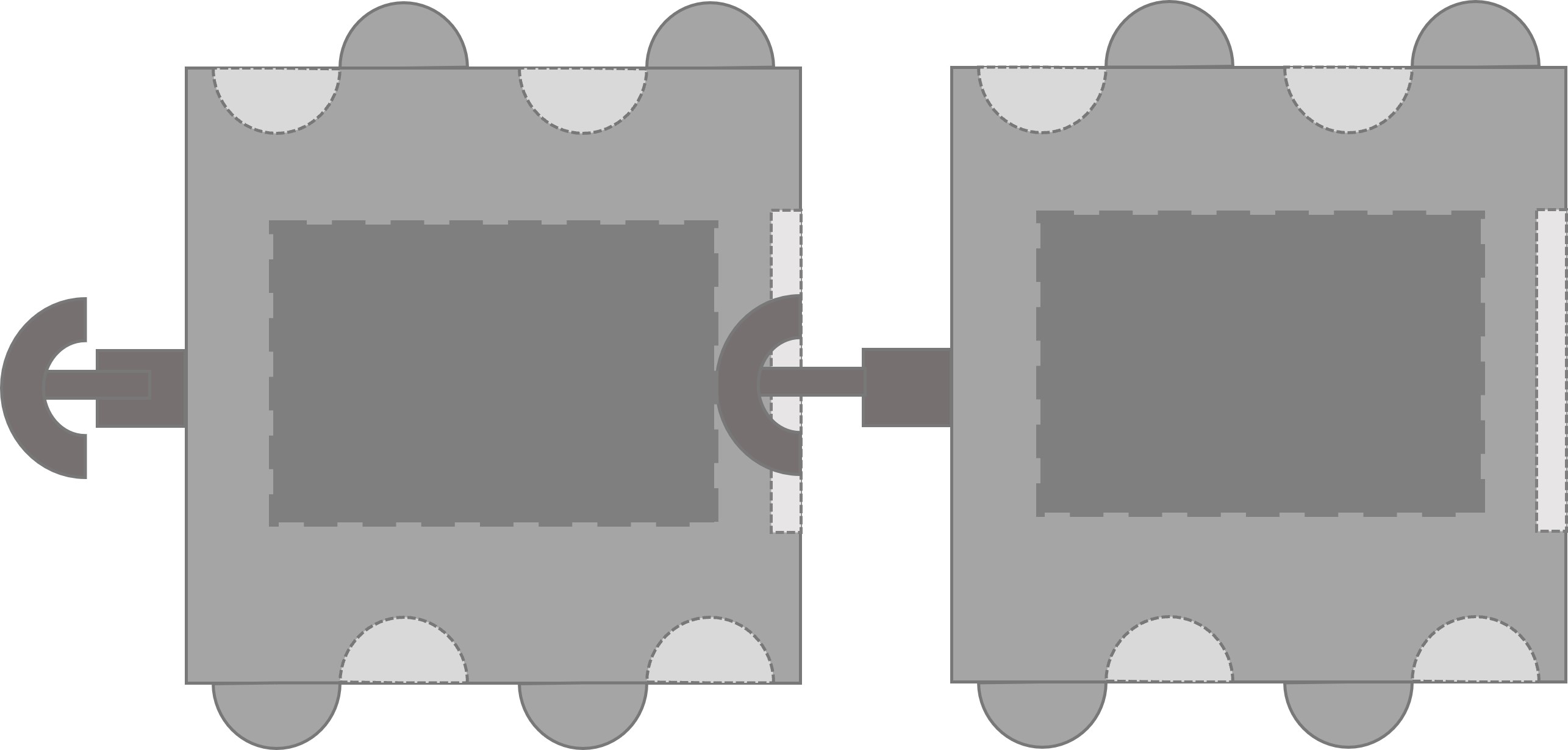}
    \caption{}
    \label{fig:anchor_couple_extend}
\end{subfigure}\hfill
\begin{subfigure}{0.32\textwidth}
    \centering
    \includegraphics[width=\textwidth]{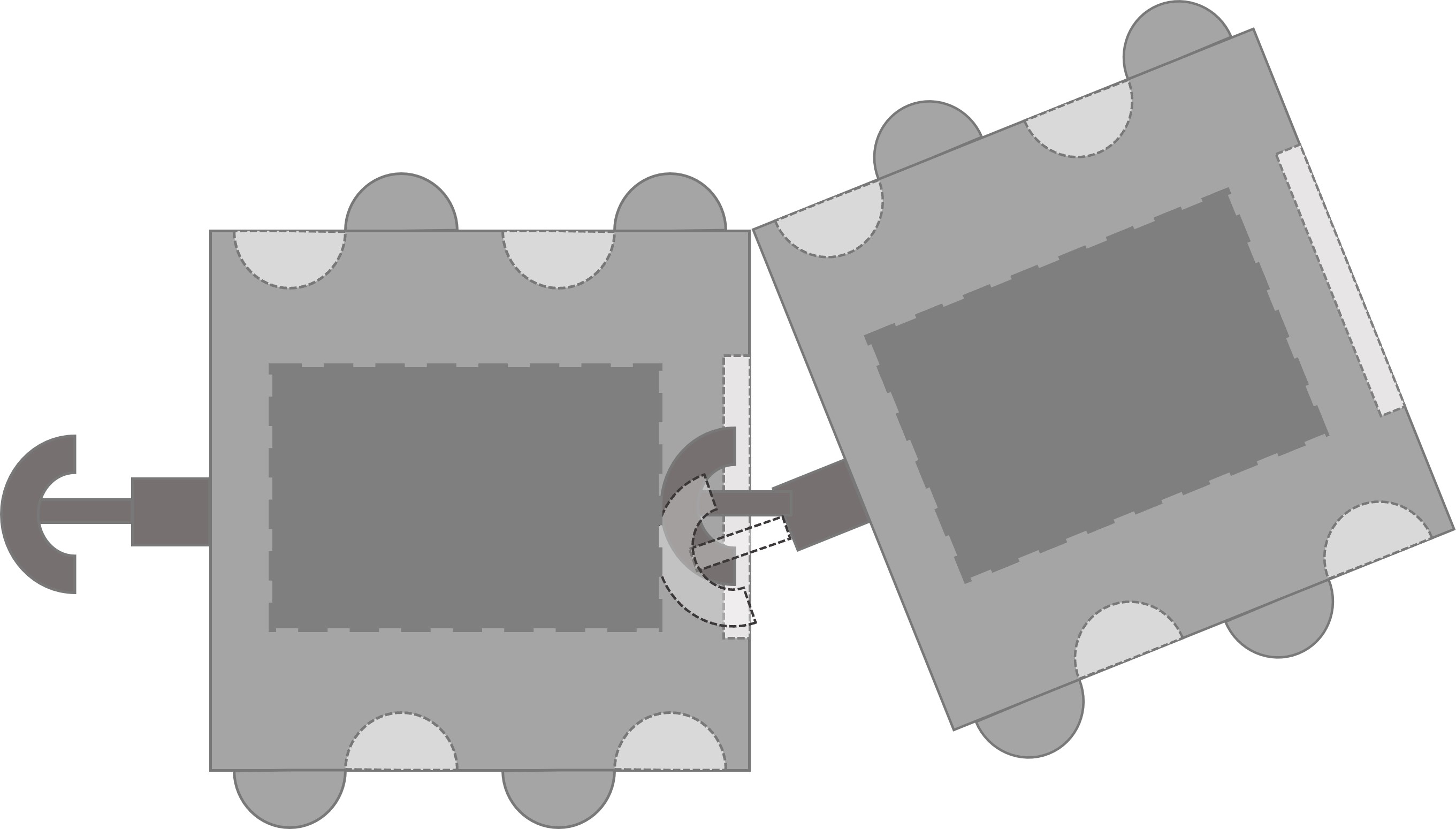}
    \caption{}
    \label{fig:anchor_couple_rotate}
\end{subfigure}\hfill
\begin{subfigure}{0.28\textwidth}
    \centering
    \includegraphics[width=\textwidth]{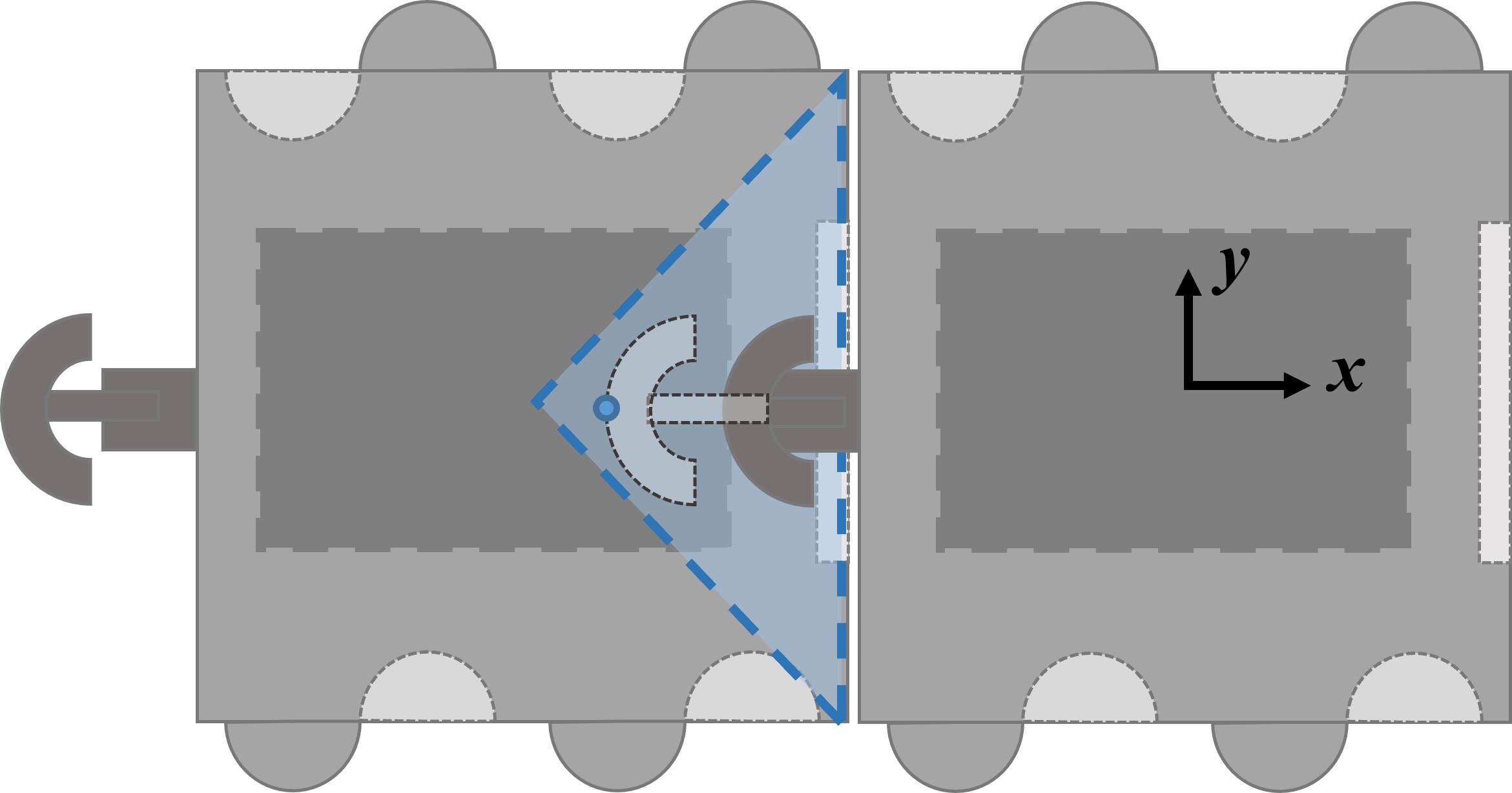}
    \caption{}
    \label{fig:anchor_couple_region}
\end{subfigure}
\caption{(a) Two robots coupled together with an extended anchor. The dark gray area in the middle of the robot is the battery box that extends to both the upper slits and the lower opening. This limits the anchor state and enforces the anchor tip to sit and lock with the opening. (b) One robot rotates as the anchor is coupled. The white shaded anchor shows a projection of the real anchor if the anchor joint is not compliant. (c) The blue shaded area shows the constraint of anchor head location during coupling.}
\vspace{-1pt}
\label{fig:anchor_couple_example}
\end{figure*}

With a passive coupling mechanism, it is essential to maintain the connection between robots by constraining the relative motion inside an assembly. In \cite{yi2022configuration}, a pair of connection points are constrained to always be within a small distance threshold to avoid accidental decoupling. This highly restricts the motion of the robots when they are assembled. With our anchor design, the floating joint, as shown in Figure~\ref{fig:anchor_holder}, provides flexibility in the rotation and translation when robots are coupled together. However, since the connection is still passive, we enforce a relaxed polygon constraint for the anchors to maintain the connection. 

As shown in Figure~\ref{fig:anchor_couple_extend}, when two robots couple, the anchor head (shown as light gray) of one robot sits inside the body of the other robot. The battery box inside the robot will block the motion of the anchor head and enforce the anchor tip to sit and lock with the striped opening shown in Figure~\ref{fig:assembly_in}. Note that, in the actual hardware system, there is some room for the anchor to move around. We define the anchor position as shown in Figure~\ref{fig:anchor_couple_extend} as the \textit{zero position} when the anchor heading angle aligns with the robot. With the floating joint between the anchor and its holder, the robot can move back and forth within a small distance and rotate within a small angle range. In the state shown in Figure~\ref{fig:anchor_couple_rotate}, the anchor of one robot sits in the opening of another robot while the robots rotate. We can model this state with a projected anchor zero position (shown in light gray). The floating joint is compliant with the robot's motion. Thus, instead of modeling the floating joint status and the corresponding robot positions, we know that the anchor is coupled with the other robot if the projected anchor zero position lies within the shaded polygon region as shown in Figure~\ref{fig:anchor_couple_region}. This polygon constraint gives a simple way of modeling the anchor connection and can be directly incorporated into the MPC framework in Section~\ref{sec:mpc_setup}.

\subsubsection{Point within Polygon}\label{sec:point_in_polygon}
Before diving into how the polygon constraint is integrated with the robot dynamics, we first need to formulate the constraint of a point inside a polygon. The Point-in-Polygon (PIP) problem is a classic computational geometry problem. Classic solutions for the PIP problem, e.g., ray-casting or winding number algorithms, are challenging to integrate with the MPC framework due to their complexity. This section presents our mathematical derivation of a set of linear constraints for a point to be inside a convex polygon. 

\begin{figure}
\centering
\includegraphics[width=0.35\textwidth]{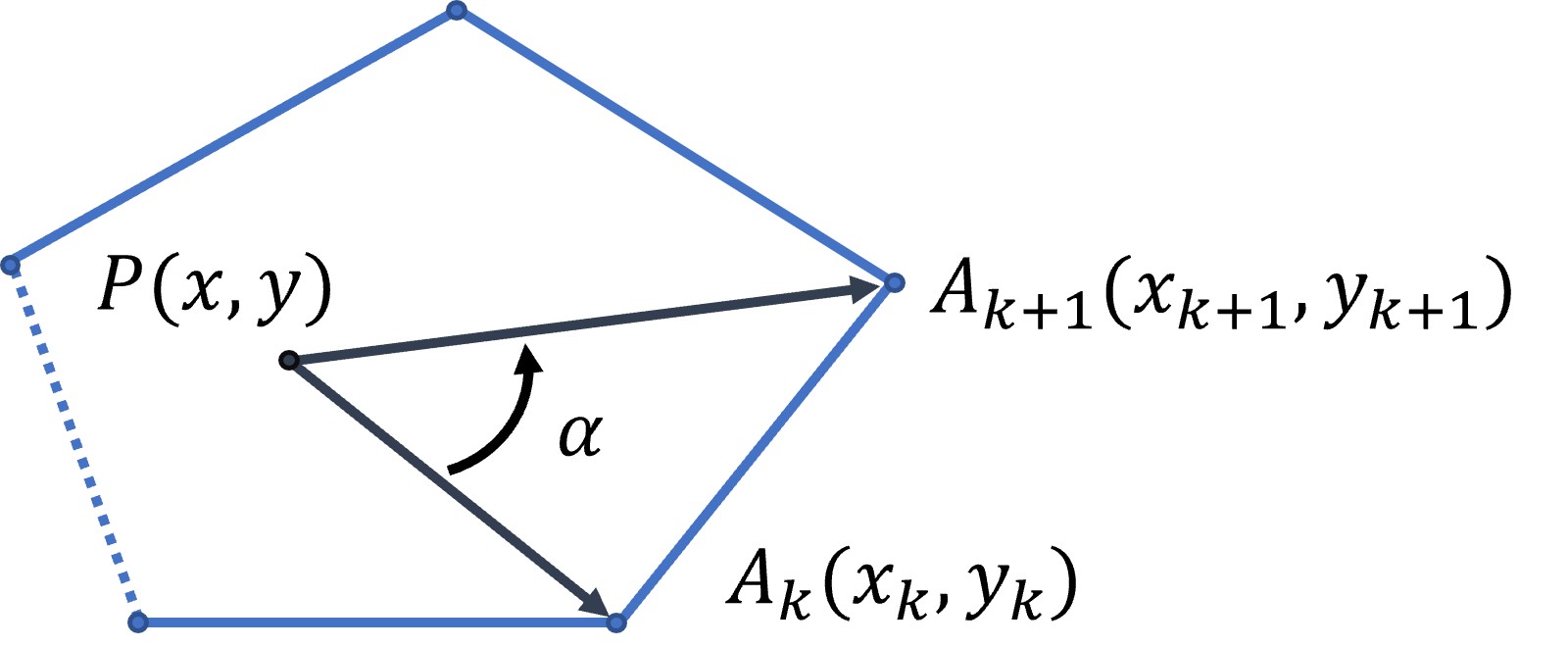}
\caption{A point inside a convex polygon.}
\label{fig:polygon_point}
\vspace{-1pt}
\end{figure}

As shown in Figure~\ref{fig:polygon_point}, a 2D convex polygon is defined by a series of points $A_1, \ldots, A_k, A_{k+1}, \ldots, A_{K}$, where $A_k \in \mathbb R ^2$ has coordinate of $(x_k, y_k)$. The numbering of the points increments counterclockwise. By definition of convex polygons, all points within the convex polygon lie on one side of each line segment $A_k A_{k+1}$. In our case of the numbering increments counterclockwise, all points within the convex polygon lie on the left-hand side of $\overrightarrow {A_k A_{k+1}}$. Consider a point $P \in \mathbb R ^2$ with a coordinate of $(x, y)$ inside the convex polygon, point $P$ should lie on the left-hand side of all $\overrightarrow {A_k A_{k+1}}$, where $k = 1, \ldots, K$, $A_{K+1} = A_1$. With the right-hand rule of cross-product, this is equivalent to 
\begin{align*}
\overrightarrow {P A_k} \times \overrightarrow {P A_{k+1}} \geq 0.
\end{align*}
Expanding this equation gives
\begin{align}
\begin{bmatrix}
    x_k - x \\ y_k - y
\end{bmatrix} \times \begin{bmatrix}
    x_{k+1} - x \\ y_{k+1} - y
\end{bmatrix} &\geq 0 \nonumber\\
(x_k - x)(y_{k+1} - y) - (y_k - y)(x_{k+1} - x) &\geq 0 \nonumber\\
\begin{bmatrix}
y_{k+1} - y_k & -(x_{k+1} - x_k)
\end{bmatrix}\left(\begin{bmatrix}x \\ y\end{bmatrix} - \begin{bmatrix} x_k \\ y_k
\end{bmatrix} \right) &\leq 0 \label{eq:polygon_linear}
\end{align}
This Equation~\eqref{eq:polygon_linear} gives a linear constraint for a point $P=(x,y)$ inside a convex polygon defined by points $A_1, \ldots, A_{K}$, which can be integrated into the MPC framework. To simplify the notation, we denote this series constraints in Equation~\eqref{eq:polygon_linear} as $pip(P| A_1, \ldots, A_{K}) \leq 0$, where the numbering of points $A_k$, $k = 1, \ldots, K$ increments counterclockwise.

\subsection{Model Predictive Control Setup}\label{sec:mpc_setup}

With the non-linear unicycle dynamics, Model Predictive Control plans based on the given non-linear dynamics and constraints and minimize a given cost function. Given the different target behaviors of the robot swarm, we may define different cost functions for the MPC. We consider the following behaviors: 1) align a set of given connection pairs; 2) go to a specific goal location; 3) fit a set of given velocity commands. 

\subsubsection{Connection Pair Alignment}
Following the same definition as in \cite{yi2022configuration}, we consider a connection pair alignment behavior where the goal is to align a set of connection pair points $\mathcal{C}_{active} = \{(C_i, C_j), \ldots\}$, $i, j \in \{1, \ldots, N\}$, where $C_i$, $C_j$ are connection points on robot $i$ and $j$ respectively. Each connection point frame $C_i$ is defined by its relative transformation $g_{B_i C_i} \in SE(2)$ to the robot body frame $B_i$. The homogeneous transformation from the Spatial (world) frame to robot frame $B_i$ is defined as
\begin{equation}\label{eq:robot_frames}
g_{S B_i} = \begin{bmatrix}
R(\theta_i) & p_i \\
0 & 1
\end{bmatrix},
g_{B_i C_i} = \begin{bmatrix}
R(\theta_{C_i}) & p_{C_i} \\
0 & 1
\end{bmatrix} \in SE(2)
\end{equation}
where $R(\theta) \in SO(2)$ is the rotation matrix corresponding to the angle $\theta$ and $p_i, p_{C_i} \in \mathbb{R}^2$ is the position vector of the frame. In our setup, $p_{C_i}$ is a predefined constant vector for each connection point. Thus, the cost of performing connection pair alignment is
\begin{align}
J_{c}(\mathcal{C}_{active}) = \sum_{(C_i, C_j) \in \mathcal{C}_{active}} \left( \Delta p^{\intercal} W_p \Delta p
+ W_{\theta}\tan^2\frac{1}{2}\Delta \theta \right)
\end{align}
where $\Delta p = R(\theta_i)p_{C_i} + p_i - (R(\theta_j)p_{C_j} + p_j)$, $\Delta \theta = \theta_i + \theta_{C_i} - (\theta_j + \theta_{C_j})$, $W_p \in \mathbb{R}^{2\times2}$ is a diagonal weight matrix for the positions, and $W_{\theta} \in \mathbb{R}$ is a scalar weight for the angle difference. To minimize angle differences, one common method is to take the angle difference directly, which gives undesirable values when the angle is around $-\pi$ or $\pi$. Another popular method is taking a modulo with $2\pi$ on the angle difference. However, this is not differentiable and significantly influences the performance of the optimization. Using quaternions instead of Euler angles is also a common approach, but it creates a huge computation overhead. Since we are in $SE(2)$ and the angle difference of $\Delta \theta$ and $\Delta \theta + 2n\pi$, $n \in \mathbb{N}$ are the same, we wrap the angle difference with a $\tan$ function that has a period of $2\pi$. 

\subsubsection{Go to goal}
Given a set of goals $\mathcal{G} = \{g_i, \ldots\}$, $g_i \in \mathbb{R}^2$, $i \in \{1, \ldots, N\}$, the cost of the robots going to the goal set is
\begin{equation}\label{eq:goal_cost}
J_g = \sum_{g_i \in \mathcal{G}} (p_i - g_i)^{\intercal}W_{g}(p_i - g_i)
\end{equation}
where $W_g \in \mathbb{R}^{2\times2}$ is a diagonal weight matrix for the goals, and $p_i$ is defined in Equation~\eqref{eq:robot_frames}. Note that here we do not verify if there are conflicting goals or not. The user or external planner is responsible for validating the goal set.

\subsubsection{Fit Command Velocity}
Fitting a given command velocity is a simple but essential part of a number of behaviors. For example, an assembly can quickly navigate around the environment with a fixed velocity. One robot can also wiggle with a slight forward/backward velocity bias to decouple from a connected assembly. Consider a given velocity command of $[v_0^*, w_0^*, \ldots, v_N^*, w_N^*]^{\intercal}$, the cost of robots achieving the given velocities is
\begin{equation}\label{eq:go_du_cost}
J_v = [v_0 - v_0^*, \ldots, w_N - w_N^*]W_v[v_0 - v_0^*, \ldots, w_N - w_N^*]^{\intercal}
\end{equation}
where $W_v \in \mathbb{R}^{2N \times 2N}$ is a diagonal weight matrix.

\subsubsection{MPC Setup with Constraints}
Consider a set of \textit{already coupled} connection pairs in the system $\mathcal{C}_{conn} = \{(C_i, C_j), \ldots\}$, $i, j \in \{1, \ldots, N\}$ similar to the active connection pair set $\mathcal{C}_{active}$. Regardless of the current behavior, it is essential to maintain the already coupled pairs. For each connection pair, we can determine on which robot the corresponding anchor locates by their relative pose. Without loss of generality and to simplify notations, here we assume the anchor of the connection pair $(C_i, C_j)$ lies on robot $i$. By utilizing the polygon constraint in Equation~\eqref{eq:polygon_linear}, we maintain each connection pair by adding the constraint $pip(C_i | R_j, C_j^r, C_j^l)$, where $C_j^r$ and $C_j^l$ denote the front right corner and the front left corner of the robot $j$ respectively. 

The MPC computation time depends highly on the number of variables and constraints. Our MPC runs in real-time at every iteration with a fixed time horizon $H_m$. However, the constraints are only essential for the first few time steps since the optimization recomputes at every time step. Thus, we incorporate the constraints only in a shorter time horizon $H_c \leq H_m$. At the time $t$, the MPC formulation is shown as follows:
\begin{align}
\min_{x, u}\ & W_{f}J(t+H_m|t) + W_c \sum_{k=0}^{H_c} J_c(\mathcal{C}_{conn}, t+k|t) \nonumber \\
&\quad\qquad + W_m\sum_{k=0}^{H_m - 1} J(t+k|t) + W_{s}u^{\intercal}u\label{eq:mpc_cost}\\
\text{s.t. }& x(t|t) = x(t) \label{eq:mpc_init_constr} \\
& x(t+k+1|t) = x(t+k|t) \nonumber\\
&\quad\qquad\quad\qquad\quad + f\left(x(t+k|t), u(t+k|t)\right)\Delta t \label{eq:mpc_dynamics_constr}\\ 
&\quad\qquad \text{for } k = 0, \ldots, H_m-1 \nonumber\\
& pip(C_{i}(t+k|t)| R_{j}(t+k|t), C_{j}^r(t+k|t), C_{j}^l(t+k|t)) \label{eq:mpc_pip_constr}\\
&\quad\qquad \text{for } k = 1, \ldots, H_c, (C_i, C_j) \in \mathcal{C}_{conn} \nonumber \\
&x(t+k|t) \in \mathcal{X}, u(t+k|t) \in \mathcal{U},\ k = 0, \ldots, H_m \label{eq:mpc_limits}
\end{align}
At each time instance $t$, we compute the above MPC formulation and execute the first control input $u(t|t)$. The cost function in Equation~\eqref{eq:mpc_cost} includes the cost of the target behavior from the behavior set $J(t+k|t) \in \mathcal{J} = \{J_c, J_g, J_v\}$, the weight of the terminal cost $W_{f}$ and stage cost $W_m$ for this behavior, a small cost with a weight of $W_c$ for maintaining the already connected pairs $\mathcal{C}_{conn}$, and a smoothness cost with a weight of $W_s$ for the control signal $u$. The optimization has four sets of constraints. The initial constraint in Equation~\eqref{eq:mpc_init_constr} is a hard state constraint so that the optimization starts from the current robots' states. Equation~\eqref{eq:mpc_dynamics_constr} is the constraints for dynamics update with the unicycle model in Equation~\eqref{eq:single_dynamics}. As mentioned previously in Section~\ref{sec:robot_dynamics}, we utilize the first-order Euler integration instead of a more accurate RK4 method. Due to our short time horizon $H_m$ and the simple dynamics model, the Euler method gives comparable results as RK4 but with less computation time. The connection pair constraints are incorporated in Equation~\eqref{eq:mpc_pip_constr} as introduced in section~\ref{sec:point_in_polygon}. Note that this constraint is only required within the constraint horizon $H_c \leq H_m$ to speed up the computation of this optimization further. We also incorporate actuation constraints in Equation~\eqref{eq:mpc_limits}. Acceleration constraint set $\mathcal{U} = \{u|u_{\min} \leq u \leq u_{\max}\}$. The velocity limits have a similar ``butterfly" shaped pattern as in \cite{yi2022configuration}, where the robot cannot stably rotate in place with a zero linear velocity, i.e., a high angular velocity comes with a high linear velocity. A controller based on only the current time instance can enforce the control signal to lie in the positive half-plane given target linear velocity is positive, and vice versa. However, this conditioned statement cannot be directly incorporated into the MPC framework as a constraint. Therefore, we define the feasible velocities in each state as $\mathcal{X} = \{x| v \leq |\frac{w_{\max}}{v_{\max}}w|\}$. This constrains the robot from not rotating in place in this MPC-based setup to further prevent the anchor from decoupling during the process. 

With this MPC framework, we can obtain the locally optimal control signal $u$ for each robot at every time instance. We then introduce when the MPC is computed and how it integrates with the current simulation and hardware system.

\subsection{Connection Pairs and Execution}
We first generate a list of connection pairs $\mathcal{C}_{goal} = \{(C_i, C_j), \ldots\}$ to couple based on a given target configuration and the distance-induced graph. Different from \cite{yi2022configuration}, which only requires the alignment of individual points, our anchor design requires a sequential coupling process. Thus, we augment each connection pair $(C_i, C_j)$ as in Algorithm~\ref{alg:augment_cp}. For each connection pair $(C_i, C_j)$, where $C_i$ is a connection point on robot $i$, we initialize its status to be \textit{decoupled}. Once the head of the anchor on one robot aligns with the opening on the other robot, the status will become \textit{head\_aligned}. Once the anchor is fully inserted, the status is \textit{head\_inserted}. Since the anchor-based connection locates only on the front and back, and the side connections are the knob-hold connection, we can define the type of connection based on the position of $C_i$ and $C_j$. Unlike the knob-hole connection, the anchor-opening connection is not symmetric. Thus, we take note of which robot the anchor is on for this given connection pair and define the corresponding anchor head position - the $C_{i}'$ point is defined at $p_{C_{i}'} = p_{C_{i}} + [-l, 0]$ if the anchor locates on robot $i$, and vice versa for $C_{j}'$, where $l$ is the length of the anchor.
\begin{algorithm}
\caption{Augment Connection Pairs}
\label{alg:augment_cp}
    \begin{algorithmic}[1]
    \Input{$\mathcal{C} = \{(C_i, C_j), \ldots\}$: set of connection pairs}
    \Output{$\tilde{\mathcal{C}}$: augmented set of connection pairs}
    \Initialize{$\tilde{\mathcal{C}}$=\{\}}
    \Function{augmentPairs}{$\mathcal{C}$}
    \For{$(C_i, C_j)$ in $\mathcal{C}$}
    \State $(C_i, C_j)$.status = \textit{decoupled}
    \State $(C_i, C_j)$.type = \textit{anchor} or \textit{knob}
    \State $(C_i, C_j)$.anchor\_index = getAnchorIndex$(C_i, C_j)$
    \State $(C_i, C_j)$.head = ($C_{i}', C_{j}'$) based on anchor\_index
    \State $\tilde{\mathcal{C}}$.append($(C_i, C_j)$)
    \EndFor
    \State \Return{$\tilde{\mathcal{C}}$}
    \EndFunction
    \end{algorithmic}
\end{algorithm}

During the execution of the connection pair list, we check if the status of each connection pair should be updated or not. As shown in Algorithm~\ref{alg:update_cp}, based on the current robot poses, we decide the connection status by checking if the connection point, or the anchor head, lies within the polygon formed by the other robot within a small margin $\epsilon$. The lists of active and connected pairs are also updated accordingly.
\begin{algorithm}[h]
\caption{Update Connection Pair Lists}
\label{alg:update_cp}
    \begin{algorithmic}[1]
    \Input{$\mathcal{C}_{conn}$: connected pairs, $\mathcal{C}_{active}$: active pairs, $\tilde{\mathcal{C}}$: augmented pair list, $\epsilon$: threshold}
    \Output{$\mathcal{C}_{conn}$: connected pairs, $\mathcal{C}_{active}$: active pairs}
    \Function{updatePairs}{$\mathcal{C}_{conn}$, $\mathcal{C}_{active}$, $\tilde{\mathcal{C}}$, $\epsilon$}
    \For{$(C_i, C_j)$ in $\mathcal{C}_{active}$}
    \State $a = (C_i, C_j)$.anchor\_index
    \If {$(C_i, C_j)$.status is \textit{decoupled}}
    \If{$(C_i, C_j)$.head in polygon($R_{\neg a}$, $\epsilon$)}
    \State $(C_i, C_j)$.status $\leftarrow$ \textit{head\_aligned}
    \State continue
    \EndIf
    \EndIf
    \If {$(C_i, C_j)$.status is \textit{head\_aligned}}
    \If{$C_i$ in polygon($R_{j}$, $\epsilon$)}
    \State $(C_i, C_j)$.status $\leftarrow$ \textit{head\_inserted}
    \State continue
    \EndIf
    \If{$(C_i, C_j)$.head not in polygon($R_{\sim a}$, $\epsilon$)}
    \State $(C_i, C_j)$.status $\leftarrow$ \textit{decoupled}
    \EndIf
    \EndIf
    \If {$(C_i, C_j)$.status is \textit{head\_inserted}}
    \State $\mathcal{C}_{active}$.remove$(C_i, C_j)$
    \State $\mathcal{C}_{conn}$.append$(C_i, C_j)$
    \EndIf
    \EndFor
    \State \Return{$\mathcal{C}_{conn}$, $\mathcal{C}_{active}$}
    \EndFunction
    \end{algorithmic}
\end{algorithm}
The connection pair alignment algorithm for the anchor-opening type of connection is shown in Algorithm~\ref{alg:connect_cp}. We obtain the goal connection pairs based on a distance-induced graph from the initial robot poses and store it for future use. We then augment the pairs as in Algorithm~\ref{alg:augment_cp}. We then find the non-conflicting pairs to execute, which are then assigned to the active list $\mathcal{C}_{active}$. We subsequently compute the control signal $u$ from the MPC introduced in Equation~\eqref{eq:mpc_cost} and update the connection pair status. The control signal acts directly on the robots in simulation. Details of the hardware system can be found in Section~\ref{sec:hardware}.
\begin{algorithm}[h]
\caption{Align Connection Pairs}
\label{alg:connect_cp}
    \begin{algorithmic}[1]
    \Input{$T$: target configuration, $x$: robot states at time $t$}
    \Output{$u$: control input}
    \Initialize{$\mathcal{C}_{conn}$=\{\}, $\mathcal{C}_{active}$=\{\}}
    \Function{alignConnectionPairs}{$T$, $x$}
    \State $\mathcal{C}_{goal} \leftarrow$ assignConnectionPairs($T$, $x$) if not assigned
    \State $\tilde{\mathcal{C}} \leftarrow$ augmentPairs($\mathcal{C}_{goal}$)
    \State $\mathcal{C}_{active} \leftarrow$ assignActivePair($\tilde{\mathcal{C}}$, $x$) without conflict
    \State $u \leftarrow$ MPC($x$, $\mathcal{C}_{conn}$, $\mathcal{C}_{active}$)
    \State $\mathcal{C}_{conn}$, $\mathcal{C}_{active} \leftarrow$ updatePairs($\mathcal{C}_{conn}$, $\mathcal{C}_{active}$, $\tilde{\mathcal{C}}$, $\epsilon$)
    \State \Return{$u$}
    \EndFunction
    \end{algorithmic}
\end{algorithm}

\section{Results and Experiments}\label{sec:result}
In this section, we first show the characterization of the 3D-printed anchors. The force profile of pushing the anchor front and back is presented. This shows that with the pushing force of each individual robot, the anchor can be pushed into the opening of another robot while it cannot be directly pulled out from the opening. In Section~\ref{sec:simulation}, we simulate the soft anchor with this force profile and test our MPC framework with the coupling behavior. In Section~\ref{sec:hardware}, we discuss in detail the structure of our hardware system and show a sequence of screenshots of robots coupling, decoupling, and going down off a test stage as a chain.

\subsection{Characterization of Anchors}\label{sec:characterization}

\begin{figure}
    \centering
    \includegraphics[width=0.4\textwidth]{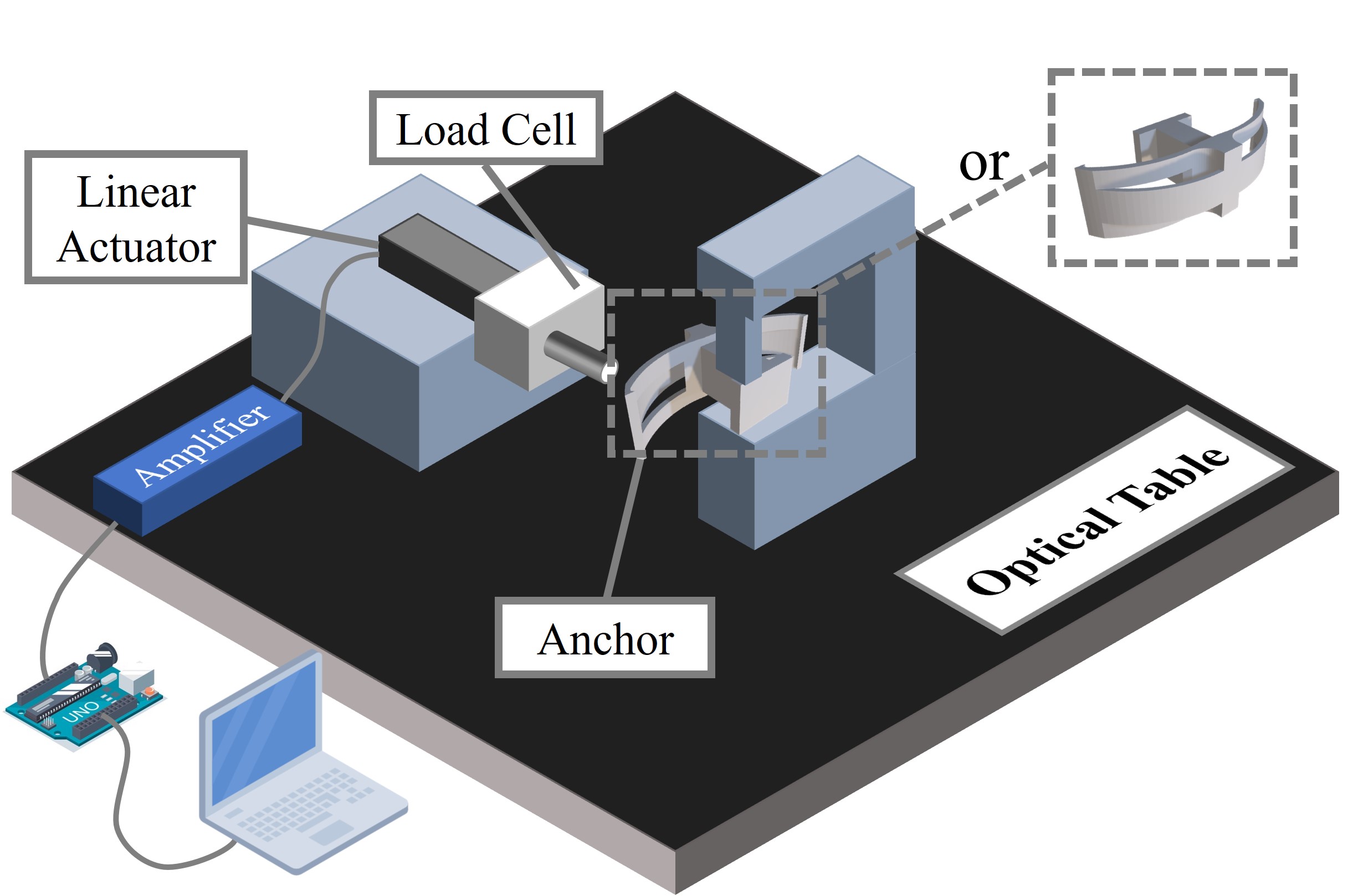}
    \caption{The force experiment setup to measure forces for pushing the anchor in the front or back depending on the positioning.}
    \label{fig:force_setup}
    \vspace{-10pt}
\end{figure}

To obtain the force profile of the 3D-printed anchor, we set up our experiment as in Figure~\ref{fig:force_setup}. A linear actuator with a distance encoder is secured on an optical table. A load cell of 500 g is attached to the linear actuator with a 3D-printed PLA holder.  An anchor is secured on the other side of the optical table. We collected data on pushing the anchor's beam from the front and back side, which resembles pushing the anchor into an opening as Figure~\ref{fig:assembly_out}, and pulling an anchor when it is locked in the opening as in Figure~\ref{fig:assembly_in}. During the experiment, an Arduino Uno sends position commands with an increment of 0.1 mm to the linear actuator. The signal from the load cell goes through an amplifier and is passed to the Arduino. Five data points are collected at each position, and an average force reading is saved. The experiments are conducted five times on three separately printed anchors respectively.  

\begin{figure}[tbp]
\begin{subfigure}{0.24\textwidth}
\centering
\includegraphics[width=\textwidth]{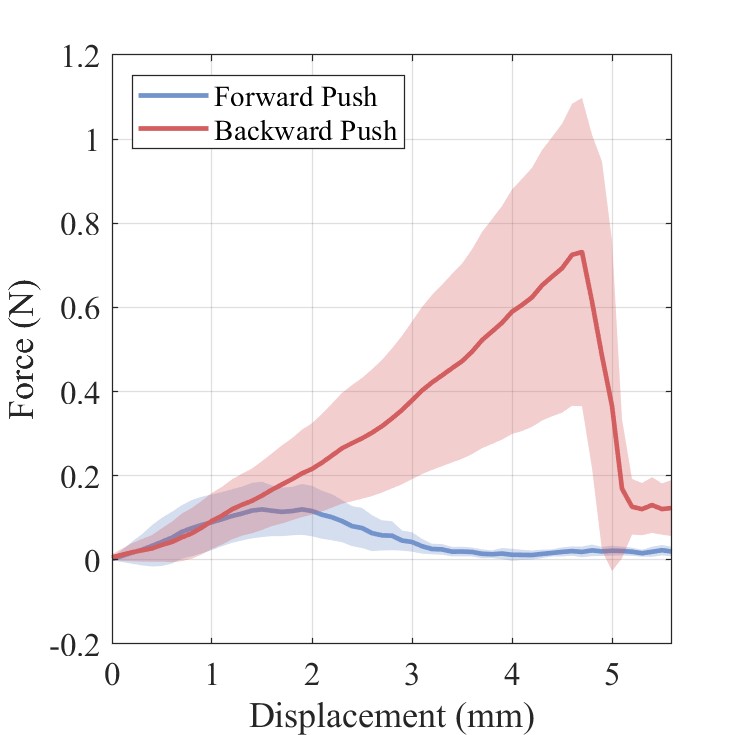}
\caption{}
\label{fig:anchor_force}
\end{subfigure}
\begin{subfigure}{0.24\textwidth}
\centering
\includegraphics[width=\textwidth]{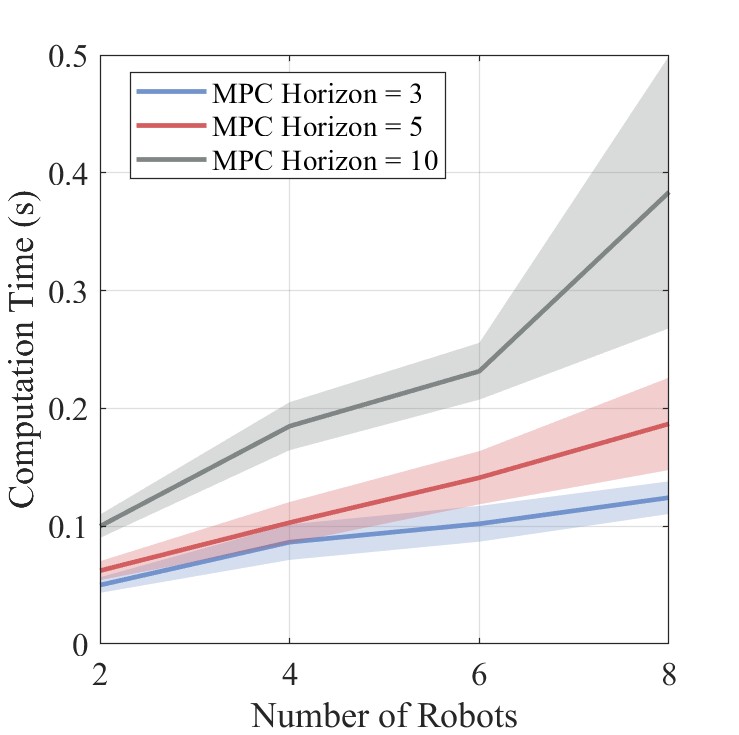}
\caption{}
\label{fig:time_vs_num_diffMPCHorizon_sumSetup}
\end{subfigure}
\caption{(a) Forces of pushing the anchor from front and back. (b) Computation time versus the number of robots with different MPC horizons. No inter-robot constraints.}
\vspace{-10pt}
\end{figure}

The result of the experiment is shown in Figure~\ref{fig:anchor_force}. When performing a forward push, which resembles the anchor being pushed into the opening of another robot as in Figure~\ref{fig:assembly_out}, the maximum force is less than 0.2 N. When performing a backward push, the average force needed for a 4 mm displacement is 0.6 N. The force drops at around 4.7 mm because of slippage due to the geometry of the anchor, which does not happen on the actual robot. The displacement needed for an anchor to be pulled out is above 3 mm. With the same setup as in Figure~\ref{fig:force_setup}, we measured that the pushing force of a robot at maximum speed has an average of 0.5 N. This will allow the anchor to easily couple, and hard to decouple with a direct pulling force. To measure the holding force when the anchor locks the connection as in Figure~\ref{fig:assembly_lock}, we put weights on the robot with the anchor locked. The anchor can hold a load of up to 500 g, which is the weight of seven robots.


\subsection{Simulation Experiments}\label{sec:simulation}
\begin{figure}
\centering
\begin{subfigure}{0.2\textwidth}
\includegraphics[width=\textwidth]{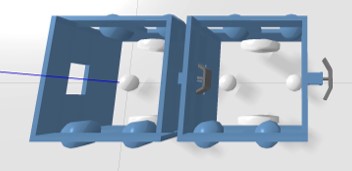}
\caption{}
\label{fig:sim_push}
\end{subfigure}
\begin{subfigure}{0.2\textwidth}
\includegraphics[width=\textwidth]{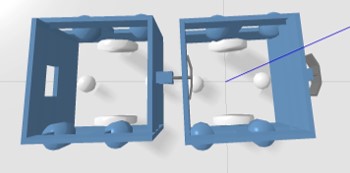}
\caption{}
\label{fig:sim_pull}
\end{subfigure}
\caption{Simulation of (a) Anchor pushed into the opening of another robot (forward push); (b) One robot pulling the anchor when it is inserted (backward push).}
\label{fig:sim_anchor}
\vspace{-10pt}
\end{figure}
\begin{figure*}[tbp]
\centering
\begin{subfigure}{0.95\textwidth}
    \centering
\includegraphics[width=\textwidth]{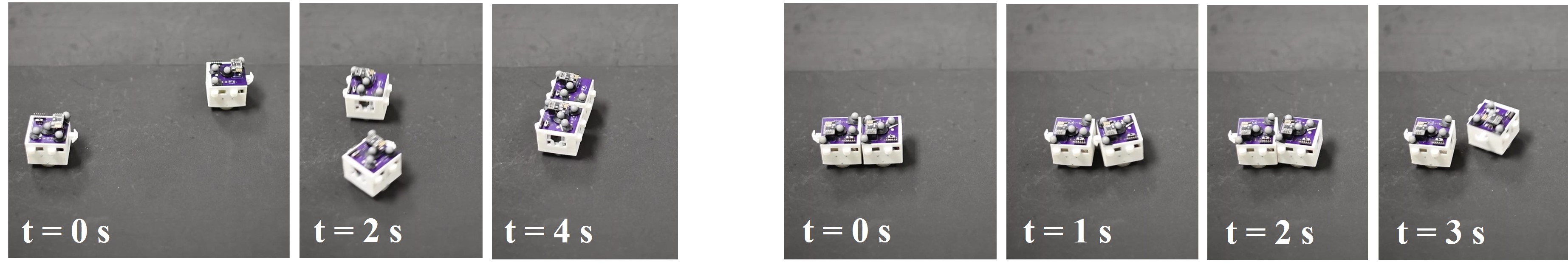}
\caption{}
\label{fig:couple_and_decouple}
\end{subfigure}
\begin{subfigure}{0.95\textwidth}
    \centering
\includegraphics[width=\textwidth]{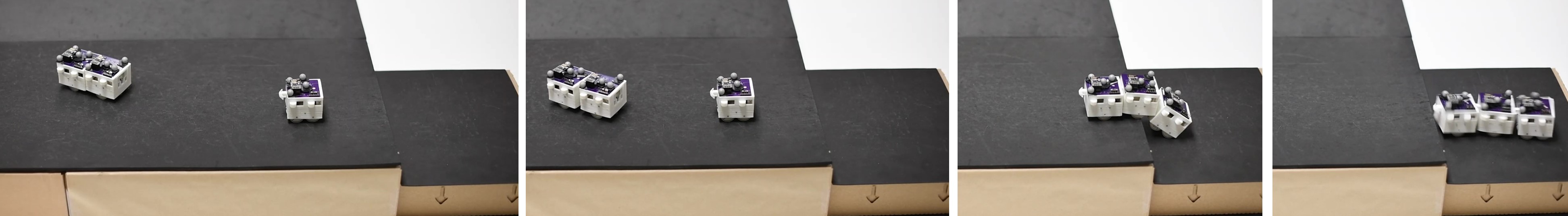}
\caption{}
\label{fig:connect_godown}
\end{subfigure}
\caption{Screenshots of (a) Left: Two robots couple with each other; Right: Two coupled robots decouple by wiggling. (b) Two coupled non-pilot robots move towards the third pilot robot, couple to form an assembly and go down a stair.}
\label{fig:hardware_screenshot}
\vspace{-8pt}
\end{figure*}
We tested our robots with anchor and opening in the Bullet \cite{coumans2021} simulation. The robot body is generated as a urdf file with a list of primitive shapes, i.e. box, cylinder, and sphere, which is experimentally more stable and accurate for collision checking compared with a custom mesh file. The robot body, anchor holder, and wheels are generated as one body, while the anchor is kept as a separate body. Since Bullet does not allow a floating joint, the anchor is placed inside the anchor holder without a joint. This multi-body setup enables stable collision checking between the anchor, anchor holder, and the opening from another robot. The soft anchor is simulated as a rigid three-bar linkage where the center bar is fixed to the anchor base, and the left and right bars are connected to the center bar with a rotational joint. The torque of the rotational joint will always try to bring the two side linkages to their resting position, with a maximum force limit corresponding to the current displacement, as shown in Figure~\ref{fig:anchor_force}. The maximum torque from the wheels also has the same limit as the actual hardware system. Figure~\ref{fig:sim_anchor} shows the simulation of the robot and its anchor. In Figure~\ref{fig:sim_push}, the robot is pushing its anchor into the opening of another robot. The side linkage bends accordingly. In Figure~\ref{fig:sim_pull}, one robot is pulling its anchor when it is inserted. The beams of the anchor are compliant but cannot be pulled out due to the limitation of the wheel torque. These behaviors resemble the actual hardware system with soft anchors.

\begin{figure}
\centering
\includegraphics[width=0.47\textwidth]{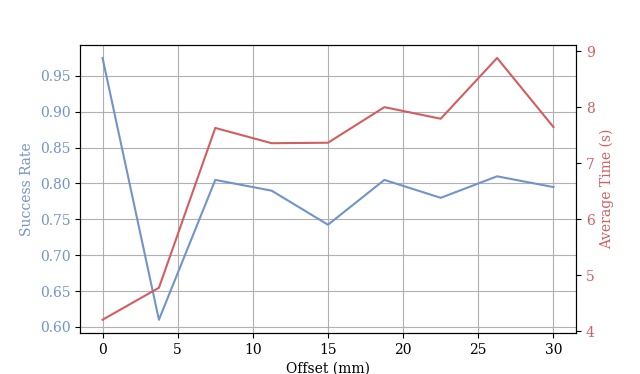}
\caption{The success rate and average time of two robots coupling with a given offset distance.}
\vspace{-10pt}
\label{fig:couple_success_rate}
\end{figure}

We tested our algorithm on a desktop with Intel Xeon 3.40 GHz CPU. The MPC framework is written in Python, interfaced with CasADi \cite{Andersson2019} with the non-linear optimization solver \textit{ip-opt} \cite{wachter2005on}. The target behavior is to couple as a line with 2, 4, 6, and 8 robots, and the computation time of the optimization is shown in Figure~\ref{fig:time_vs_num_diffMPCHorizon_sumSetup}. The MPC horizon $H_m$ tested are 3, 5, and 10, with the same constraint horizon $H_c = 3$. Each time step $\Delta t=0.1$ s. We can see that the computation time increases as the number of robot increase, and with a larger MPC horizon, the computation time increases significantly. We also tested the success rate and average time for two robots to couple. We initialize two robots with a displacement of 65 $mm$ in the x direction, and an offset ranging from 0 to 30 $mm$ in the y direction. Since the coupling controller will run non-stop until the robots are coupled, we regard the coupling behavior to be successful if coupled within one minute. Each experiment is repeated 100 times, and the results are shown in Figure~\ref{fig:couple_success_rate}. We may see that with a perfect alignment, i.e. zero offset, robots have the highest success rate of 97.5\%, and the shortest coupling time. While a small offset of less than 8 $mm$ significantly decreases the success rate, the robots have more space to adjust their alignment with our controller, given a larger offset of greater than 8 $mm$. Since the robots need to start from coupled status for the experiments on decoupling behavior, only one set of experiments is conducted with fixed starting positions. With 100 trails, the success rate is 99\%, with an average completion time of 7.02 seconds.

\subsection{Hardware Experiments}\label{sec:hardware}


The hardware system runs in ROS \cite{ros}. The MPC planner node runs at 10 Hz. It takes in the current robot poses, computes the optimal linear and angular acceleration, and outputs the corresponding linear and angular velocity. The optimization running in this MPC node is exactly the same as in the Bullet simulation while having a different interface on the hardware system. A separate high-frequency PID controller node runs at 30 Hz. The controller node takes the Vicon pose, calculates the instant linear and angular velocity within a small time window, and computes the wheel motor values based on the given command velocity from the MPC node. 

A series of screenshots from our experiment video is shown in Figure~\ref{fig:hardware_screenshot}. Figure~\ref{fig:couple_and_decouple} shows two robot couple with each other and decouple by having one robot wiggle with a linear velocity bias. Figure~\ref{fig:connect_godown} shows an assembly of two coupled robots connected with a pilot robot. They successfully formed a chain and went down a stair. More videos of the hardware experiments can be found in our supplementary video material.

\section{Conclusions and Future Works}
In this paper, we presented our soft anchor design that has an asymmetric force profile when pushing from the front and back. This makes the robots easy to couple while having a sufficient holding force for the connection to carry high loads. When two robots are coupled, the anchor acts as a compliant joint that provides a limited rotation and translation. We also presented our MPC framework with polygon constraints to model the coupling behavior. The polygon constraints capture the geometry of connections. Thus, the assembly is flexible to rotate, move around the environment, and form 3D structures while maintaining their coupling status. The side knobs of the robot enable the robots to form rigid structures. With this setup, the robot swarm can form functional structures in an unstructured environment, e.g., bridges across gaps and ropes down a stair. 

The current framework has limitations as follows. The centralized MPC is computationally expensive for more than eight robots. By moving the MPC to decentralized based on assembly segments and performing the computation in parallel, we can improve the scalability of the system. The current anchor system is designed for robots in centimeter-scale. Further experiments will be done to study how the anchor system scales with different sizes of robots. If the robot can be equipped with onboard sensors and creates a map of the unstructured environment, we can also computationally determine the required assembly shape based on the environment structure.

\section*{Acknowledgments}
We thank Xinyu Wang for the initial anchor design and testing. This work is supported by grant AFOSR FA9550-18-0251.

\bibliographystyle{plainnat}
\bibliography{references}

\end{document}